\pdfoutput=1

\documentclass[11pt]{article}

\usepackage{EMNLP2023}

\usepackage{times}
\usepackage{latexsym}

\usepackage[T1]{fontenc}

\usepackage[utf8]{inputenc}

\usepackage{microtype}

\usepackage{inconsolata}
\usepackage{graphicx}
\usepackage{multirow}
\usepackage{booktabs}
\usepackage{tabularx}
\usepackage{enumitem}
\setlist[itemize]{nosep,leftmargin=*}
\usepackage{amsmath}

\title{Multi-level Contrastive Learning for Script-based Character Understanding}

\author{\textbf{Dawei Li\textsuperscript{1}, Hengyuan Zhang\textsuperscript{2}, Yanran Li\textsuperscript{3}}, Shiping Yang\textsuperscript{4} \\
  \textsuperscript{1}Halicioğlu Data Science Institute, University of California, San Diego \\
  \textsuperscript{2}Shenzhen International Graduate School, Tsinghua University \\
  \textsuperscript{3}Independent Researcher \\
  \textsuperscript{4}School of Computer Science, Beijing University of Posts and Telecommunications\\ 
  \texttt{dal034@ucsd.edu,} \quad \texttt{zhang-hy22@mails.tsinghua.edu.cn,} \\
  \texttt{yanranli.summer@gmail.com,} \quad \texttt{yangshipingnlp@gmail.com,} \\
}

\begin{document}
\maketitle
\begin{abstract}
In this work, we tackle the scenario of understanding characters in scripts, which aims to learn the characters' personalities and identities from their utterances. We begin by analyzing several challenges in this scenario, and then propose a multi-level contrastive learning framework to capture characters' global information in a fine-grained manner. To validate the proposed framework, we conduct extensive experiments on three character understanding sub-tasks by comparing with strong pre-trained language models, including SpanBERT, Longformer, BigBird and ChatGPT-3.5. Experimental results demonstrate that our method improves the performances by a considerable margin.
Through further in-depth analysis, we show the effectiveness of our method in addressing the challenges and provide more hints on the scenario of character understanding.
We will open-source our work in this \href{https://github.com/David-Li0406/Script-based-Character-Understanding}{URL}.

\end{abstract}

\section{Introduction}

As one essential element in stories, character comprehension is a popular research topic in literary, psychological and educational research~\cite{mckee1997story,currie2009narrative,paris2003assessing,bower1978experiments}. To fully understand characters, individuals must empathize with characters based on personal experiences~\cite{gernsbacher1998automatically}, construct profiles according to characters' identities, and inference about characters' future actions~\cite{fiske1979imaging,mead1990representation}.

\begin{table}[]\small
\centering
\begin{tabular}{p{0.6in}p{1.0in}p{0.9in}}
\hline
Character    & Sheldon                                                                                                                                                      & Jennifer                                                                                                                                       \\ \hline
Story Title  & TBBT                                                                                                                                          & The Test                                                                                                                                       \\ \hline
Dataset      & TVSHOWGUESS                                                                                                            & ROCStories                                                                                                                                     \\ \hline
Text Length  & 528832                                                                                                                                                       & 41                                                                                                                                             \\ \hline
Character's Related Text & Sheldon: " ... we take on Koothrappaliand his dog. Really give ourselves a challenge." & Jennifer has a big exam tomorrow. ... Jennifer felt bittersweet about it. \\ \hline
\end{tabular}
\caption{Comparison between a script from TVSHOWGUESS~\cite{sang2022tvshowguess} and a narrative from ROCStories~\cite{mostafazadeh2016corpus}.}
\label{Comparison}
\end{table}

According to the data modality and format, character comprehension can be categorized into several classes~\cite{sang2022survey}. 
In this work, we focus on character understanding in scripts~\cite{chen2016character,sang2022tvshowguess}. 
Scripts are written text for plays, movies, or broadcasts~\cite{onions1966oxford}. Typically, scripts are often structured with several text fields, including scene description, conversation, transition and summary~\cite{Saha_Movie_Script_Database_2021}.

Although pre-trained language models (PLMs) have demonstrated their effectiveness in language and vision research fields~\cite{qiu2020pre,min2021recent}, script-based character understanding is yet a hard task, as shown in our experiments. Here we highlight two challenges. 
The first one is \textbf{text type}.
As scripts mainly consist of conversations between different characters, at the core of script-based character understanding is conversation understanding.
Especially, scripts often involve multi-party conversations where multiple characters talk and interact with each other in a single scene. Considering other common issues in conversation understanding, it is non-trivial for PLMs to comprehend characters based on fine-grained conversation information~\cite{li2021self,ma2022structural,li2022c3kg,tu2022misc}. 
The other challenge of applying PLMs to script-based character understanding is \textbf{text length}. Table~\ref{Comparison} shows a comparison between a script from TVSHOWGUESS~\cite{sang2022tvshowguess} and a short story from ROCStories~\cite{mostafazadeh2016corpus}. 
Typically, scripts are very long with even billion of words~\cite{chen2016character}, and in turn character information are distributed globally throughout the entire script~\cite{bai2021joint,inoue2022learning}. However, PLMs are ineffective in capturing such global information due to the sensitiveness of context modeling~\cite{liu2019roberta,joshi2020spanbert} and the limitation of input length~\cite{dai2019transformer,beltagy2020longformer}. 

To address the aforementioned challenges, we propose a multi-level contrastive learning framework and capture both fine-grained and global information using two devised contrastive losses. For fine-grained character information, we build a \textbf{summary-conversation contrastive loss} by comparing character representations from different sources. Specifically, we leverage two text fields in scripts, i.e., summary and conversation, and then extract character representations from the corresponding field. The representations of the same character are then treated as the postive pairs, while those of different characters are negative pairs. To model the global information, we also propose a novel \textbf{cross-sample contrastive loss} as inspired by~\cite{bai2021joint,inoue2022learning}.
By aligning the same character's representation in different samples, the model overcomes the input length limitation and learns the global information of each character.
To validate the effectiveness of our framework, we benchmark the performances of several PLMs, including SpanBERT, Longformer, BigBird, and ChatGPT-3.5, 
on three widely-adopted character understanding tasks. 


In general, our contributions are as follows:
\begin{itemize}
    \item We identify two critical challenges for character understanding in scripts and propose a multi-level contrastive learning framework to address them.
    \item Through extensive experiments, we demonstrate the effectiveness of our method across multiple datasets and downstream tasks.
    \item With further analysis, we provide some insights into script-based character understanding. All codes will be open-sourced for future research.
\end{itemize}

\section{Related Work}

\subsection{Character Understanding}
Character understanding has long been the subject of considerable interest and scrutiny. Some early works propose to extract keywords as characters' features from movies~\cite{bamman2013learning} and novels~\cite{flekova2015personality}. Other works attempt to learn the relationship between characters in both supervised~\cite{massey2015annotating,kim2019frowning} and unsupervised ways~\cite{krishnan2014you,iyyer2016feuding}.

Recently, more challenging tasks in character understanding have emerged.~\citet{chen2016character} benchmark the character linking and coreference resolution tasks on TV show scripts.~\citet{brahman2021let} collect dataset with storybooks and their summaries, and define the character description generation and character identification tasks. \citet{sang2022tvshowguess} extend the character guessing task into a multi-character scenario on TV show scripts. Additionally, some works attempt to combine traditional self-supervised learning methods~\cite{mikolov2013distributed} with language models~\cite{liu2019roberta} to learn contextual character embeddings and apply them in downstream tasks~\cite{azab2019representing,inoue2022learning}.

In this work, we focus on character understanding tasks in scripts. 
While some works benchmark summary-based tasks in narratives~\cite{chen2021summscreen,brahman2021let}, we are the first to leverage script summaries as auxiliary data and learn fine-grained and global character representations in a novel way.

\begin{figure*}[!t]
    \centering
    \includegraphics[width=16cm]{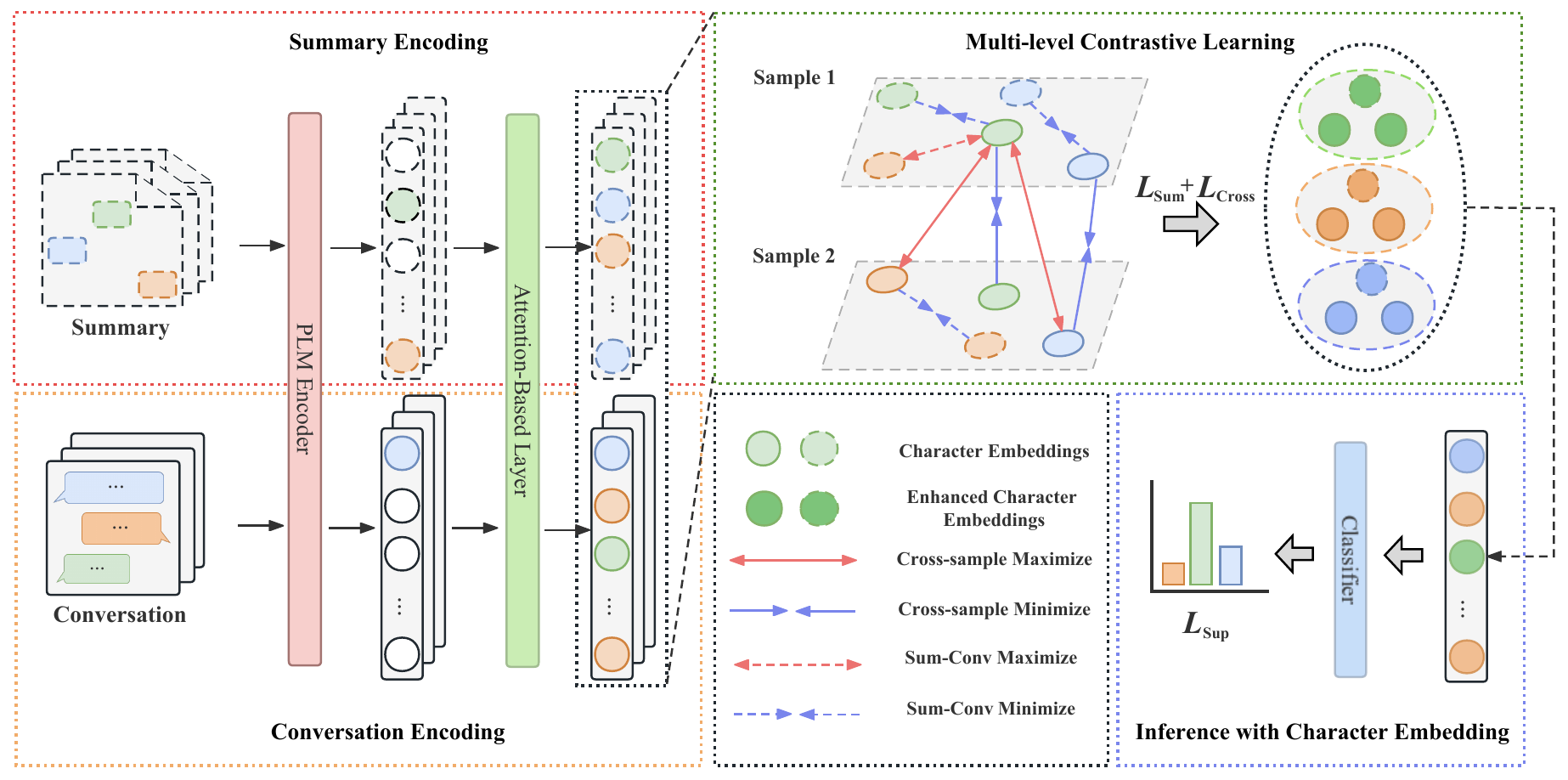}
    \caption{The overview pipeline of our method. Each color represents a character entity or embedding. The conversation and summary encoding parts correspond to Section~\ref{Character Representation in Conversation} and~\ref{Character Representation in Summary} respectively. The multi-level contrastive learning part corresponds to Section~\ref{Multi-level Contrastive Learning}. The inference with character embedding part corresponds to Section~\ref{Two-stage Training}.}
    \label{overview}
\end{figure*}

\subsection{Contrastive Learning}
In recent years, contrastive learning is widely used in various NLP tasks~\cite{zhang2022contrastive}, including sentence representation~\cite{gao2021simcse,kim2021self}, machine translation~\cite{pan2021contrastive,vamvas2021contrastive}, text generation~\cite{lee2020contrastive,shu2021logic,zhang2022fine}, and etc.
Literatures in multimodal research field adopt contrastive learning for vision-language model training, constructing positive pairs with images and their corresponding captions~\cite{li2020oscar,radford2021learning,yang2022unified}.
In our work, we also regard characters in summaries and conversations as two different views of the same target and align them for a better representation.

Moreover, some works aim to construct positive pairs in global manners. Both~\citet{qin2020erica} and~\citet{hogan2022fine} conduct document-level contrastive learning in the relation extraction task to align the representation of the same entity and relation. \citet{pan2021contrastive} propose an aligned augmentation method that generates more positive sentence pairs in different languages to improve translation performances in non-English directions. Similarly, \citet{qin2022gl} acquire multilingual views of the same utterance from bi-lingual dictionaries. Following this line of research, we propose the cross-sample contrastive learning in addition to the in-sample contrastive loss to learn character representations globally.


\section{Preliminaries}

Generally, character understanding tasks require the model to predict character's information given a segment of text.
For script-based character understanding, the provided texts often consist of conversations within scripts.
In this work, we also leverage script summaries as an additional source.
We provide detailed examples in Appendix~\ref{Example of Script-based Character Understanding Task}.

In practice, the model first generates character's embeddings $e$ in the representation learning step. Subsequently, a feed-forward network {\rm FFN} is often adopted as the classifier with the cross-entropy loss:

\begin{equation}
    p = Softmax({\rm FFN}(e))
\end{equation}

\begin{equation}
    L_{Sup} = -\frac{1}{N} \sum_{i=1}^{N} y_i \log(p)
\end{equation}


\section{Method}


Our work presents a multi-level contrastive learning framework for character representation learning. Firstly, we follow a general encoding process to obtain character representations from conversations and summaries. 
Then, we describe two novel contrastive losses to capture fine-grained and global information at both in-sample and cross-sample levels.
Finally, we propose a two-stage training paradigm that applies different losses in different learning stages.
Figure~\ref{overview} illustrates an overview pipeline of our method.

\subsection{Character Representation in Conversation}
\label{Character Representation in Conversation}
To obtain character representations from the conversation field in the scripts, we first concatenate each utterance~\cite{joshi2020spanbert,beltagy2020longformer} and utilize a pre-trained language model ${\rm PLM}$\footnote{Without loss of generalization, we adopt several PLMs in experiments.} to produce the encoding of the whole text $\mathbf{H}$:

\begin{equation}
    \mathbf{H} = {\rm PLM}(u_1;u_2;,...;u_T)
\end{equation}


Then, the character embeddings $e_1,e_2,...e_n$ are extracted from the contextual encoding $\mathbf{H}$.
After that, we follow previous works~\cite{bai2021joint,sang2022tvshowguess} and use an attention-based layer to share the character-level information among each embedding\footnote{We provide further details in Appendix~\ref{Details of Character Embedding Generation}}:

\begin{equation}
    e_1,...e_n = {\rm Extract}(\mathbf{H})
\end{equation}

\begin{equation}
    e_1,...e_n = {\rm Attention}(e_1,...e_n)
\end{equation}


However, the conversations in the scripts are complex and thus the character embeddings solely based on the conversations are often insufficient for fine-grained character understanding. 

\subsection{Character Representation in Summary}
\label{Character Representation in Summary}

To supply more information, we leverage scripts' summaries as auxiliary data and apply contrastive learning to capture the character intricacies. 

Similar with conversation encoding, given a summary $S$ contains a group of character mentions $\{cm_1^s, cm_2^s,..., cm_n^s\}$, we also encode the whole summary and extract the character representations:
\begin{align}
    \mathbf{H}_s &= {\rm PLM}(S) \\
      e_i^s &= t_{start_i} + t_{end_i}, 1<=i<=n
\end{align}
where $t_{start_i}$ and $t_{end_i}$ are the first and last tokens of the $i_{th}$ character mention $cm_i^s$ in the summary.

After that, we follow~\cite{bai2021joint} and use a mention-level self-attention (MLSA) layer \footnote{It is a transformer encoder layer with $B$ repeated block. Please refer to~\citet{bai2021joint} for more details.} to gather information for each character embedding:

\begin{equation}
    e_1^s, ...,e_n^s = {\rm MLSA}(e_1^s, ...,e_n^s)
\end{equation}
and the last layer's output $e_i^s$ is treated as the character's representation from the summary.

\subsection{Multi-level Contrastive Learning}
\label{Multi-level Contrastive Learning}

To enhance the character representations learned from the conversation and the summary, we develop a novel multi-level contrastive learning to capture both fine-grained and global information.

\subsubsection{Summary-conversation Contrastive Learning}

At the local in-sample level, we develop a summary-conversation contrastive loss to align representations of the same character. This gives the model an additional perspective on character representation and encourages it to find a general space where different representations of the same character are closer. Concretely, the loss function for the summary-conversation contrastive learning is:


\begin{equation}
    L_{Sum} = \sum_{i=1}^P -log\frac{exp^{{\rm sim}(e_{c_i},e_{c_i}^s)}/\tau}{\sum_{j=1}^P exp^{{\rm sim}(e_{c_i},e_{c_j}^s)}/\tau}
\end{equation}
where $c_i$ denotes the $i_{th}$ character, and $P$ here is the number of characters that appear in both scripts and summaries. Also, $\tau$ is a temperature hyper-parameter, and $\rm sim(,)$ stands for the similarity function\footnote{Here we use Cosine similarity.}. 
Note that in samples where conversation and summary contain multiple representations of character $c_i$, we randomly select one as $e_{c_i}$ and $e_{c_i}^s$, respectively.

By applying the summary-conversation contrastive loss, we are able to learn fine-grained character representations from both summary and conversation texts. 

\subsubsection{Cross-sample Contrastive Learning}

In addition to fine-grained information, global-level information is also crucial for character representation learning~\cite{bai2021joint,inoue2022learning}. 
To this end, we also propose a cross-sample contrastive learning to align the same character representation in different samples within a batch:



\begin{equation}
\label{Cross-sample contrastive loss function}
    L_{Cross} = \sum_{i=1}^K-log\frac{exp^{{\rm sim}(e_{c_i}^1,e_{c_i}^2)}/\tau}{\sum_{j=1}^K exp^{{\rm sim}(e_{c_i}^1,e_{c_j}^2)}/\tau}
\end{equation}

\begin{equation}
\label{Gaurantee}
    SI(e_{c_i}^1) \neq SI(e_{c_i}^2)
\end{equation}
where $SI(e)$ means the sample index of the character representation $e$\footnote{$e$ generally represents any character embedding.}. 
When there are multiple representations of one given character in a batch, we randomly select two from them. 
For cross-sample learning, we impose a constraint that restricts $e_{c_i}^1$ and $e_{c_i}^2$ to originate from different samples.
$K$ is the number of characters appearing in at least two different samples within a batch. 
To this end, the cross-sample contrastive loss forces the model to utilize global information in a batch and thus obtain a comprehensive understanding of the characters.






\subsection{Two-stage Training}
\label{Two-stage Training}

To fully train the model, we further propose a two-stage training paradigm to apply different losses in different learning stages.

Concretely, in the first stage, we combine the two contrastive losses with the supervised loss together, and post-train the pre-trained language model. 
The supervised loss serves as a guidance to facilitate the contrastive learning, and stabilize the training at the very beginning. The total loss of the first stage is:

\begin{equation}
\label{task ratio define}
    L_{Total} = \lambda*L_{Sup} + \alpha*L_{Sum} + \beta*L_{Cross}
\end{equation}
where $\lambda,\alpha,\beta$ are hyper-parameters of task ratios, and we will analyze their effects in Section~\ref{Analysis on Hyper-Parameter}.
After the first stage, only the supervised loss is kept to train the model in the second stage. This makes the model concentrate on the downstream supervision signals. 

\section{Experiments Setup}

\subsection{Tasks and Datasets}
\label{Tasks and Datasets}
We evaluate the proposed method on three character understanding tasks, i.e., coreference resolution~\cite{chen2016character}, character linking~\cite{chen2016character}, and character guessing~\cite{sang2022tvshowguess}.

\noindent\textbf{Coreference Resolution} Given a conversation in scripts that contains multiple utterances and $n$ character mention entity $c_1, c_2, ..., c_n$ within it, the objective of the coreference resolution task is to assemble all mention entities that refer to the same character in a cluster.

\noindent\textbf{Character Linking} The input of the character linking task is the same as coreference resolution. 
Unlike coreference resolution, the goal of character linking is to accurately classify each mention entity to the character in a pre-defined character set $Z = \{z_1, z_2, ..., z_m \}$.

\noindent\textbf{Character Guessing} Distinct from previous tasks, the character guessing task focuses on identifying the speaker for each utterance in scripts. 
In this task, each utterance within a scene is segmented and fed into the model. 
The speaker's name preceding each utterance is masked and replaced with a special token. 
The same speaker within a scene is represented by the same special token. 
The objective of the character guessing task is to predict the identity of the speaker for each special token.

\noindent\textbf{Datasets} We choose two TV show datasets to conduct experiments. For coreference resolution and character linking, we use the latest released version of the Character Identification dataset\footnote{\url{https://github.com/emorynlp/character-identification}}. 
For character guessing, we adopt the TVSHOWGUESS dataset\footnote{\url{https://github.com/YisiSang/TVSHOWGUESS}} to conduct experiments. We follow all the training, development, and testing separation provided by the original datasets. The dataset statistics are given in Table~\ref{datasets} in Appendix.

\subsection{Baseline Models}
Following previous works, we adopt several state-of-the-art (SOTA) models in character understanding as baselines and apply the proposed framework on them. For coreference resolution and character linking, we choose \textbf{SpanBERT}~\cite{joshi2020spanbert}, a transformer-architecture pre-trained model with the contiguous random span mask strategy in the pre-training stage. 
We also adopt \textbf{C$^\textbf{2}$}, which combines coreference resolution and character linking together and achieves the SOTA performance in both two tasks. 
For character guessing, we use \textbf{BigBird}~\cite{zaheer2020big} and \textbf{Longformer}~\cite{beltagy2020longformer}, as they are specialized for long-form document input. We follow~\citet{sang2022tvshowguess} and add a character-specific attentive pooling layer upon the the model encoders and denote them as \textbf{BigBird-P} and \textbf{Longformer-P}. Notably, we also design a zero-shot and one-shot instruction prompts and evaluate \textbf{ChatGPT-3.5} (gpt-3.5-turbo) via its official API\footnote{\url{https://platform.openai.com/docs/api-reference/completions/create}} as another strong large language model baseline. 





\subsection{Evaluation Metrics}
For coreference resolution, we follow the previous works~\cite{zhou2018they,bai2021joint} and use B3, CEAF$\phi$4, and BLANC as our evaluation metrics. 
These three metrics are first proposed by the CoNNL'12 shared task~\cite{pradhan2012conll} to measure the clustering performance of the coreference resolution task. 
For character linking and character guessing, we use Macro and Micro F1 to evaluate the models' classification performances.

\begin{table*}[h]\small
\centering
\setlength{\tabcolsep}{1.3mm}{
\begin{tabular}{@{}l|ccc|ccc|ccc|c|c@{}}

\toprule
\multirow{3}{*}{MODEL} & \multicolumn{9}{c}{Coreference Resolution}                                                                                                             & \multicolumn{2}{|c}{Character Linking}           \\ \cmidrule(l){2-12} 
                       & \multicolumn{3}{c|}{B3}                           & \multicolumn{3}{c|}{CEAF$\phi$4}                       & \multicolumn{3}{c|}{BLANC}                        & \multirow{2}{*}{\centering MICRO} & \multirow{2}{*}{\centering MACRO} \\ \cmidrule(l){2-10}
                       & PREC.          & REC.           & F1             & PREC.          & REC.           & F1             & PREC.          & REC.           & F1             &                        &                        \\ \bottomrule
ChatGPT-Zero-Shot         & 63.43          & 59.51          & 61.41          & 68.39          & 64.37          & 66.32          & 80.39          & 77.74          & 78.97          & 74.7                   & 64.3                   \\
ChatGPT-One-Shot         & 66.43          & 62.54          & 64.43          & 68.47          & 64.44          & 66.40          & 82.19          & 79.40          & 80.70          & 76.2                   & 63.6                   \\ \bottomrule
SpanBERT-base         & 77.40          & 82.67*          & 79.94          & 74.69          & 67.93          & 71.15*          & 84.80*          & 89.96          & 87.20          & 85.0*                   & 78.4                   \\
SpanBERT-base (Ours)  & 79.95*          & 84.71          & 82.26          & 76.67          & 70.38          & 73.39*          & 87.44          & 91.26          & 89.26          & 86.3                   & 78.9*                   \\
SpanBERT-large        & 81.92          & 85.56          & 83.69*          & 77.85          & 74.74          & 76.25*          & 88.61*          & 91.91          & 90.20          & 87.2*                   & 82.8*                   \\
SpanBERT-large (Ours) & 83.55*          & \textbf{87.38*}          & 85.42*          & \textbf{79.83}          & 76.29          & 78.02*          & 89.18*          & 93.00          & 91.00          & \textbf{88.2*}                   & \textbf{83.7*}                   \\ \bottomrule
C$^2$-base                & 80.75          & 84.77*          & 82.71*          & 76.97          & 71.78          & 74.28          & 82.22*          & 91.52          & 89.80*          & 85.6                   & 80.4*                   \\
C$^2$-base (Ours)         & 83.35          & 85.12*          & 84.23          & 76.88*          & 74.97          & 75.91          & 90.48          & 91.85*          & 91.15          & 86.4                   & 81.1                   \\
C$^2$-large               & 84.98          & 86.92* & 85.94          & 79.63 & 78.16*          & 78.89          & 90.87*          & 93.05*          & 91.93          & 87.6*                   & 82.5*                   \\
C$^2$-large (Ours)        & \textbf{86.42} & 86.44*          & \textbf{86.24*} & 78.82          & \textbf{80.42*} & \textbf{79.61} & \textbf{91.77*} & \textbf{93.13} & \textbf{92.45*} & 88.0*          & 83.2*          \\ \bottomrule
\end{tabular}}
\caption{Automatic evaluation results on coreference resolution and character linking. The best results are in bold. We follow previous works to present the results of coreference resolution in a 2-digital decimal and the results of character linking in a 1-digital decimal. * denotes that $p \leq 0.01$ in the statistical significance test.}
\label{Coreference Resolution & Character Linking}
\end{table*}

\subsection{Implementation Details}
We employ both the base and large sizes of each model, and implement our proposed method on them. For summary-conversation contrastive loss, we use summary corpus collected by~\citet{chen2021summscreen}. We follow the hyper-parameter settings in the original papers to reproduce each baseline's result. 
We repeat each experiment 3 times and report the average scores. For ChatGPT prompts and other implementation details, please refer to Appendix~\ref{Prompt for ChatGPT} and Appendix~\ref{Detailed Training Settings}.
We will open-source all codes in this work.

\begin{table}[h]\small
\centering
\setlength{\tabcolsep}{3.6mm}{
\begin{tabular}{@{}lcc@{}}
\toprule
Model                     & \multicolumn{1}{c}{MICRO}             & \multicolumn{1}{c}{MACRO}             \\ \midrule
ChatGPT-Zero-Shot         &  48.58          & 42.17          \\
ChatGPT-One-Shot         &  51.57          & 44.05          \\ \midrule
BigBird-P-base        &   71.01                                    &       70.32*                                \\ 
BigBird-P-base (Ours) &    72.61                                   &     73.00 \\
BigBird-P-large        &   75.43*                                    &       75.24                                \\ 
BigBird-P-large (Ours) &    77.68*                                   &     76.41 \\ \midrule
Longformer-P-base         &  71.80          & 73.75          \\
Longformer-P-base (Ours)  &  73.65* &  74.22 \\ 
Longformer-P-large        &   77.58                                    &       75.92*                                \\ 
Longformer-P-large (Ours) &  \textbf{78.92*}                                     &       \textbf{76.52*} \\
\bottomrule
\end{tabular}}
\caption{Automatic evaluation results on character guessing. The best results are in bold.}
\label{Charcter Guessing}
\end{table}

\section{Results and Analysis}

\subsection{Main Results}

Table~\ref{Coreference Resolution & Character Linking} and Table~\ref{Charcter Guessing} present the automatic evaluation results on the three tasks. Surprisingly, even with specialized instruction and one-shot demonstration, ChatGPT-3.5 performs the worst among all the baselines on each task. This implies that character understanding is still hard and complex to solve for large language models. 
Among the three tasks, models perform worse on character guessing than on coreference resolution and character linking tasks. In particular, ChatGPT achieves extremely low scores of 44.05 Macro-F1 in character guessing. Since character guessing requires a deeper understanding of each character and more varied narrative comprehension skills~\cite{sang2022tvshowguess}, this suggests that \textbf{the current pre-trained models, especially LLMs, have room for improvement in tasks that require global and in-depth learning for a specific individual}.

Despite the discrepancies in model architecture and size, the proposed method brings significant improvements to each baseline model on almost every metric, except for B3 and CEAF$\phi$4 in C$^2$-large model. These results indicate the effectiveness and compatibility of our method.

\begin{table*}[t]\small
\centering
\setlength{\tabcolsep}{1.0mm}{
\begin{tabular}{@{}l|ccc|ccc|ccc|c|c@{}}
\toprule
\multirow{3}{*}{MODEL} & \multicolumn{9}{c}{Coreference Resolution}                                                                                                             & \multicolumn{2}{|c}{Character Linking}           \\ \cmidrule(l){2-12} 
                       & \multicolumn{3}{c|}{B3}                           & \multicolumn{3}{c|}{CEAF$\phi$4}                       & \multicolumn{3}{c|}{BLANC}                        & \multirow{2}{*}{\centering MICRO} & \multirow{2}{*}{\centering MACRO} \\ \cmidrule(l){2-10}
                       & PREC.          & REC.           & F1             & PREC.          & REC.           & F1             & PREC.          & REC.           & F1             &                        &                        \\ \bottomrule
SpanBERT-base                            & 77.40          & 82.67          & 79.94          & 74.69          & 67.93          & 71.15          & 84.80          & 89.96          & 87.20          & 85.0                   & 78.4                   \\
SpanBERT-base (Ours)                     & \textbf{79.95}          & \textbf{84.71}          & \textbf{82.26}          & \textbf{76.67}          & 70.38          & \textbf{73.39}          & 87.44          & \textbf{91.26}          & \textbf{89.26}          & \textbf{86.3}                   & 78.9                   \\
w/o cross-sample loss         & 79.48          & 83.06          & 81.23          & 74.72          & \textbf{71.07}          & 72.85          & \textbf{87.68}          & 90.59          & 89.08          & 86.1                   & \textbf{80.0}                   \\
w/o summary-conversation loss & 79.00          & 83.36          & 81.11          & 74.68          & 70.33          & 72.44          & 85.45          & 90.61          & 87.85          & 85.6                   & 78.8                   \\ \bottomrule
SpanBERT-large                           & 81.92          & 85.56          & 83.69          & 77.85          & 74.74          & 76.25          & 88.61          & 91.91          & 90.20          & 87.2                   & 82.8                   \\
SpanBERT-large (Ours)                    & 83.55          & \textbf{87.38}          & \textbf{85.42}          & \textbf{79.83}          & 76.29          & \textbf{78.02}          & 89.18          & \textbf{93.00}          & \textbf{91.00}          & \textbf{88.2}                   & \textbf{83.7}                   \\
w/o cross-sample loss         & \textbf{83.85}          & 86.68 & 85.24          & 79.44 & 76.37          & 77.88          & \textbf{90.68}          & 92.47          & 90.65          & 87.4                   & 83.6                   \\
w/o summary-conversation loss & 85.29 & 83.96          & 84.62 & 74.65          & \textbf{79.20} & 76.69 & 91.45 & 91.15 & 90.80 & 87.9         & 82.8          \\ \bottomrule
\end{tabular}}
\caption{Ablation study results on two contrastive losses. The experiment is conducted using character resolution and character linking.}
\label{Ablation Study}
\end{table*}

\begin{table}[h]\small
\centering
\setlength{\tabcolsep}{0.9mm}{
\begin{tabular}{@{}l|ccc|cc@{}}
\toprule
                  & \multicolumn{3}{c|}{Coreference Resolution}                                                & \multicolumn{2}{c}{Character Linking}             \\ \midrule
                  & \multicolumn{1}{c|}{B3}             & \multicolumn{1}{c|}{CEAF$\phi$4}         & BLANC          & \multicolumn{1}{c|}{MICRO}         & MACRO         \\ \midrule
C2-base           & \multicolumn{1}{c|}{82.71}          & \multicolumn{1}{c|}{74.28}          & 89.80          & \multicolumn{1}{c|}{85.6}          & 80.4          \\ \midrule
C2-base-OS & \multicolumn{1}{c|}{83.58}          & \multicolumn{1}{c|}{74.28}          & 90.63          & \multicolumn{1}{c|}{86.1}          & 80.8          \\ \midrule
C2-base-TS & \multicolumn{1}{c|}{\textbf{84.23}} & \multicolumn{1}{c|}{\textbf{75.91}} & \textbf{91.15} & \multicolumn{1}{c|}{\textbf{86.4}} & \textbf{81.1} \\ \bottomrule
\end{tabular}}
\caption{Ablation study results on two-stage learning strategy. -OS and -TS represents the one-stage training and two-stage training respectively. For one-stage training, we remove the second supervised loss-only stage and adopt the multi-task training only.}
\label{ablation study on two-stage learning}
\end{table}

\subsection{Ablation Studies}

We also conduct an ablation study to examine the contributions of the two novel contrastive losses, i.e., the cross-sample loss and summary-conversation loss. 
To implement, we select SpanBERT-base and SpanBERT-large as backbone models and implement model variants by removing one of two contrastive losses in the training phases.


Table~\ref{Ablation Study} presents the results of our ablation study on the coreference resolution and character linking tasks. Compared with the vanilla SpanBERT-base and SpanBERT-large, adding one or two contrastive losses yield better performances. Additionally, we observe that when applied separately, models with the summary-conversation loss work better than models with the cross-sample loss only. More importantly, it is evident that the models trained with both contrastive losses together outperform the models with only one loss, indicating the necessity of our multi-level contrastive framework as well as its effectiveness in addressing the two challenges, i.e., text type and text length.

We also conduct an ablation study on the two-stage learning strategy.
Table~\ref{ablation study on two-stage learning} shows the experiment results on C2-base using character linking and coreference resolution.
While the one-stage multi-task training can also improve the baseline model's performance, we found it leads to a sub-optimal result compared with that using our two-stage learning strategy.
This observation leads us to the conclusion that supervision-only fine-tuning is also very important in our method, consistently enhancing baseline models' performance.
This aligns with the findings of prior research, which advocate for task-specific fine-tuning following multi-task post-training~\cite{guan2020knowledge,han2021fine}.

\begin{table}[h]\small
\centering
\setlength{\tabcolsep}{0.8mm}{
\begin{tabular}{@{}c|c|c|c|c|c|c|c@{}}
\toprule
$\lambda$ & $\alpha$ & $\beta$ & B3 & CEAF$\phi$4 & BLANC & MICRO & MACRO \\ \midrule
--               & --              & --           & 83.69 & 76.25    & 90.20 & 87.2  & 82.8     \\ \midrule
1.0             & 0.0            & 0.0         & 83.98 & 76.12    & 90.75  & 87.8  & 82.2    \\
0.0             & 1.0            & 1.0         & 85.04 & 77.72    & 90.77  & 86.4  & 78.6    \\ \midrule
1.0             & 1.0            & 1.0         & 85.42 & 78.02    & 91.00 & 88.2  & 83.7      \\
0.5             & 1.0            & 1.0         & 85.14 & 78.15    & 91.02  & 88.4  & 83.2    \\
1.0             & 0.5            & 0.5         & 85.23 & 79.00    & 90.96  & 88.1  & 82.0    \\ \bottomrule
\end{tabular}}
\caption{Hyper-parameter analysis results on coreference resolution and character linking. For coreference resolution, we report the F1 scores of the B3, CEAF$\phi$4 and BLANC metrics.} 
\label{Hyper-parameter Analysis}
\end{table} 

\subsection{Analysis on Hyper-Parameters}
\label{Analysis on Hyper-Parameter}
The task ratio setting is also an important component of our method. 
In this section, we investigate their impacts by testing various task ratios in the first training stage. 
We employ the SpanBERT-large model and perform experiments on the coreference resolution and character linking tasks.


The results of the hyper-parameter analysis are presented in Table~\ref{Hyper-parameter Analysis}.
As defined in Equation~\ref{task ratio define}, $\lambda$, $\alpha$, and $\beta$ represent the ratios of task-specific supervised loss, summary-conversation loss, and cross-sample loss, respectively. Accordingly, the first block (Row 1) presents the vanilla SpanBERT-large performance w/o our framework, and the second block (Row 2 and Row 3) shows the model variants with only supervision loss or contrastive losses. Comparing the first and second block we can see, there is no obvious improvement when only keeping the supervised loss, a.k.a $\lambda = 1.0, \alpha = 0.0, \beta = 0.0$ in the first stage. 
Moreover, when $\lambda$ is set to $0.0$, the model trained without supervised loss also exhibits inferior performances, e.g., there is a notable decrease in Macro F1 (from 82.8 to 78.6). 
This finding supports our hypothesis that \textbf{the task-specific supervision signal plays a crucial role in guiding the two contrastive learning}.
When examining the last block (Row 4-6), we observe that the models w/ our framework under different task ratios consistently surpasses the others (except only one MARCO metric). This further demonstrates the robustness of our method on the task ratio hyper-parameter.

\begin{figure}[!t]
    \centering
    \includegraphics[width=7.5cm]{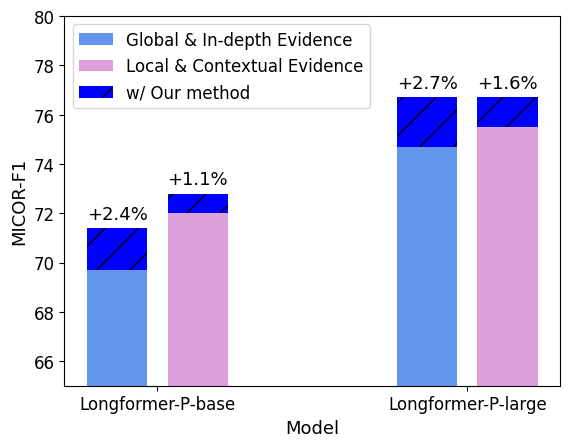}
    \caption{Evidence type analysis result.} 
    \label{Evidence Type Fig}
\end{figure}

\subsection{Resource Availability Analysis}

The proposed summary-conversation contrastive learning relies on well-organized script datasets that include a summary of each scene.
This prerequisite could potentially limit the applicability of our approach to datasets in other languages or domains.
To address this constraint, we conduct an experiment in which we replaced the manually collected summary dataset with an automatically generated one, produced by ChatGPT.
As depicted in Table~\ref{tab: Resource Availability Analysis}, our results indicate that when using the auto-generated corpus in summary-conversation contrastive learning, a significant improvement is still observed when compared to the vanilla baseline.
This discovery further validates the adaptability of our method, irrespective of whether golden or generated summaries are used.

\begin{table}[h]\small
\centering
\setlength{\tabcolsep}{0.60mm}{
\begin{tabular}{@{}l|ccc|cc@{}}
\toprule
            & \multicolumn{3}{c|}{Coreference Resolution}                                                & \multicolumn{2}{c}{Character Linking}             \\ \midrule
            & \multicolumn{1}{c|}{B3}             & \multicolumn{1}{c|}{CEAF$\phi$4}         & BLANC          & \multicolumn{1}{c|}{MICRO}         & MACRO         \\ \hline
C2-base           & \multicolumn{1}{c|}{82.71}          & \multicolumn{1}{c|}{74.28}          & 89.80          & \multicolumn{1}{c|}{85.6}          & 80.4          \\ \midrule
C2-base-LLM & \multicolumn{1}{c|}{84.14}          & \multicolumn{1}{c|}{\textbf{76.06}} & 90.89          & \multicolumn{1}{c|}{86.1}          & 80.9          \\ \midrule
C2-base-G   & \multicolumn{1}{c|}{\textbf{84.23}} & \multicolumn{1}{c|}{75.91}          & \textbf{91.15} & \multicolumn{1}{c|}{\textbf{86.4}} & \textbf{81.1} \\ \bottomrule
\end{tabular}}
\caption{Experiment results with automatically generated summarization. -LLM and -G denote the model trained on summaries generated by ChatGPT and those trained using the dataset provided by~\cite{chen2021summscreen}.}
\label{tab: Resource Availability Analysis}
\end{table}

\subsection{Breakdown to Evidence Type}

To better understand when and how our method works on each sample, we conduct an evidence type analysis on the character guessing task based on the fine-grained annotation provided by~\citet{sang2022tvshowguess}.  
To remedy the scarcity issue in the original annotations, we merge the fine-grained annotation categories into two broader categories: \emph{Global \& In-depth Evidence} and \emph{Local \& Textual Evidence}. 
More details on evidence type merging is described in Appendix~\ref{Evidence Type Merging}.

The results of evidence type analysis are presented in Figure~\ref{Evidence Type Fig}. Note that our framework works better when Local \& Textual evidence is required for character guessing than Global \& In-depth evidence. This finding aligns with our intuition that Global \& In-depth evidence is more challenging for the model to comprehend. 
It is also worth noting that our framework yields larger increases for samples requiring Global \& In-depth evidences (2.4\% and 2.7\% for the base and large size models respectively), as compared to those requiring Local \& Textual evidence (1.1\% and 1.6\% for the base and large models respectively). Based on these results, we safely conclude that \textbf{our framework is effective in facilitating character information modeling, especially for global information}.

\subsection{Visualization}

The core of our method is to learn fine-grained and global character representations. To this end, we also visualize the learned character embeddings in the character guessing task. Specifically, we use character embeddings in the test set of the ``FRIENDS'' (a subset of TVSHOWGUESS dataset) and randomly choose 6 embeddings for each character from different samples. 

\begin{figure}[!t]

    \centering
    \includegraphics[width=7.5cm]{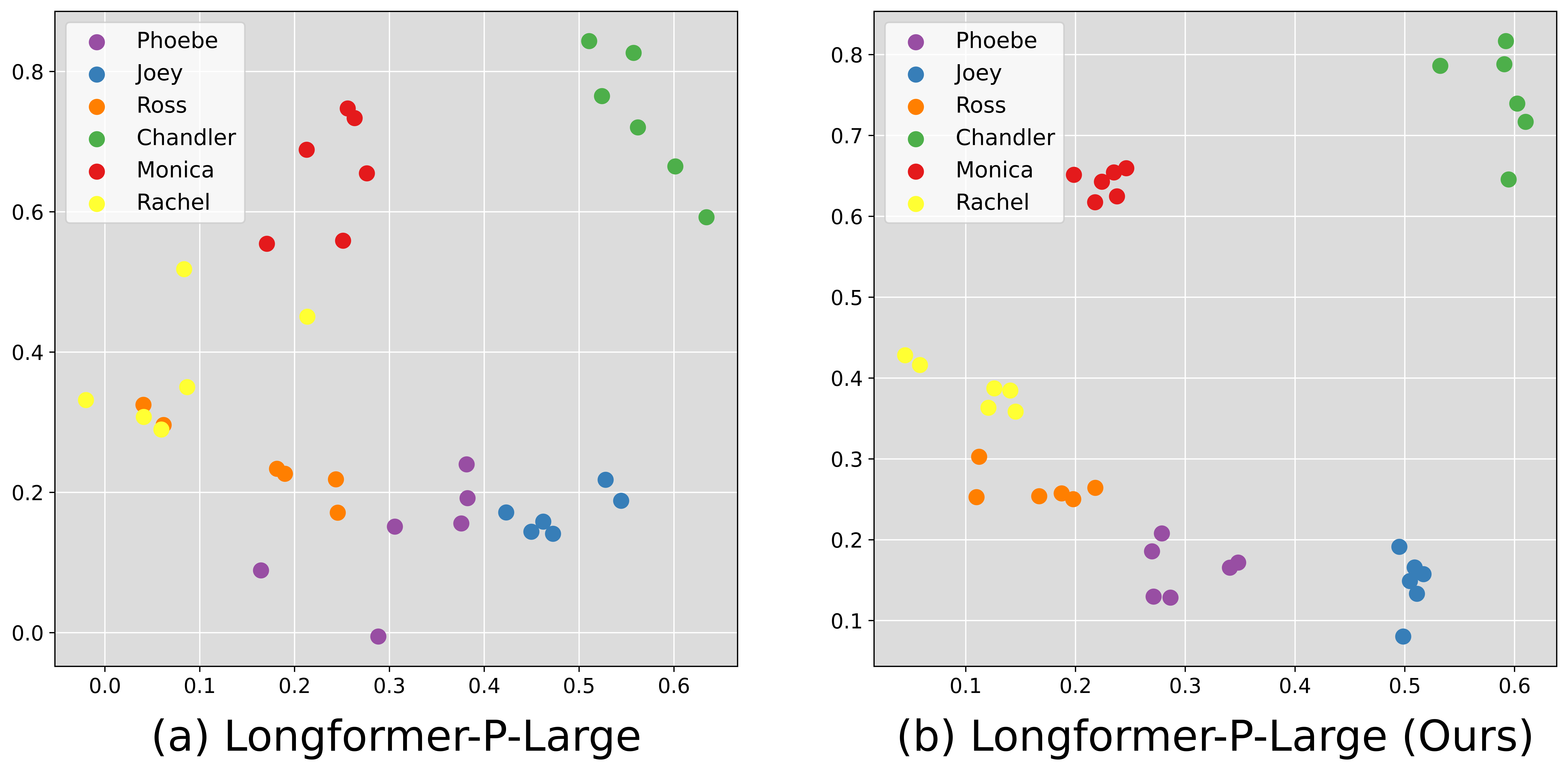}
    \caption{Character embedding visualization result.} 
    \label{Visulization Fig}
\end{figure}

Figure~\ref{Visulization Fig} shows the visualization results using T-SNE~\cite{van2008visualizing}. 
We compare the character embeddings generated by Longformer-P-Large w/ and w/o our framework. One thing to note is that without our framework, some character embeddings of Ross overlap with those of Rachel. This is because that in the TV show ``FRIENDS'', Ross and Rachel are partners and together appearing and engaging in many scenes. In contrast, this overlapping phenomenon is greatly mitigated. Overally speaking, our framework encourages the embeddings belonging to the same character exhibit a more compact clustering pattern. This finding provides a new perspective to understand the effectiveness of our proposed method in character comprehension tasks.

\subsection{Case Study}

We also choose a challenging sample from ``The Big Bang Theory'' subset of TVSHOWGUESS in the character guessing task, and analyze the predictions from Longformer-P-Large w/o and w/ our method, as well as that from ChatGPT.

As shown in Table~\ref{Case Study}, all the predictions from ChatGPT are wrong, indicating ChatGPT lacks a fine-grained understanding of each character. Besides, the only difference between the vanilla model w/ and w/o our framework is whether the speaker P1 is predicted correctly or not. 
In this case, predicting P1 is particularly challenging, as few utterances are spoken by this character. Hence, it is a must for the models to guess P1's identity using other details in the scene. By understanding the relationships between P1 and other characters, our method is able to correctly predict that P1 is Sheldon's partner, Amy. 
This demonstrates that \textbf{our method benefits the fine-grained understanding on character relationships in script-based character understanding}, e.g., character guessing tasks.

\begin{table}[h!]\small
\centering
\begin{tabular}{@{}l@{}}
\toprule
\begin{tabular}[c]{@{}l@{}}P0 : Hey, sorry about that\\ P1 : No, we're sorry. We never should have been\\ comparing relationships in the first place.\\ P2 : Why? We won. You know, I say, next, we take\\ on Koothrappali and his dog. Really give ourselves\\ a challenge.\\ P3 : I just want to say one more thing about this.\\ Just because Penny and I are very different people\\ does not mean that we're a bad couple.\\ P2 : The answer is one simple test away. \\ Hmm? You know, it's like when I thought there was\\ a possum  in my closet. Did I sit around wondering?\\ No, I sent Leonard in with a pointy stick and a bag.\\ P3 : I killed his Chewbacca slippers.\\ P0 : Let's just take the test.\\ P3 : No, no, no, I don't want to.\\ P0 : Oh, well, 'cause you know we're gonna do bad.\\ P3 : Because it doesn't matter. I don't care if we're\\ a ten or a two.\\ P2 : Or a one. A one is possible.\\ P3 : Marriage is scary. You're scared, I'm scared. But\\ it  doesn't make me not want to do it. It, it just\\ makes me want to hold your hand and do it with you.\\ P0 : Leonard.\\ P1 : It makes me so happy if you said things like that.\\ P2 : We got an eight-point-two. Trust me, you're happy.\end{tabular} \\ \midrule
ChatGPT:\emph{P0: Leonard, P1: Sheldon, P2: Penny, P3:Howard} \\ \midrule
Vanilla: \emph{P0: Penny, P1: Howard, P2: Sheldon, P3:Leonard}\\ \midrule
Ours: \emph{P0: Penny, P1: Amy, P2: Sheldon, P3:Leonard}\\ \midrule
Golden: \emph{P0: Penny, P1: Amy, P2: Sheldon, P3:Leonard}\\ \bottomrule
\end{tabular}
\caption{An example chosen from ``The Big Bang Theory'' in the character guessing task. We analyze the predictions made by ChatGPT (one-shot), Longformer-P-Large (vanilla and with our framework).}
\label{Case Study}
\end{table}

\section{Discussion about LLMs on Character Understanding}
In this section, we go deeper to discuss the unsatisfied performance when adopting the ICL of LLMs to perform character understanding tasks.
One possible reason for this is the script-based character understanding we focus on requires the model to learn the character information globally. For example, in character guessing, anonymous speakers sometimes need to be identified with some global evidence, like linguistic style and the character's relationship with others. These subtle cues are usually not included in the current sample and thus require the model to learn them globally from other samples~\cite{sang2022tvshowguess}. However, due to the fine-tuned unavailability of ICL, LLMs can only utilize local information from the current sample and limited demonstrations to make inferences. We believe this is the reason that LLMs don't perform well in our script-based character understanding scenario.
Additionally, we notice ICL also falls short in some other tasks that involve learning a domain-specific entity or individual across multiple samples, like knowledge graph completion~\cite{yao2023exploring}.

In our work, we notice while the number of demonstrations increases, the performance of LLMs shows a corresponding improvement.
It appears that augmenting the number of demonstrations in the prompt could be a potential strategy for enhancing the capabilities of LLMs in these global learning tasks.
Nonetheless, it's essential to note that incorporating an excessive number of relevant samples as demonstrations faces practical challenges, primarily due to constraints related to input length and efficiency considerations.
In the future, more efforts are needed to explore optimal ways of harnessing the ICL method of LLMs in such global learning scenarios.

\section{Conclusions}

In this work, we focus on addressing two key challenges, text length and text type in script-based character understanding. 
To overcome these challenges, we propose a novel multi-level contrastive framework that exploits in-sample and cross-sample features. The experimental results on three tasks show that our method is effective and compatible with several SOTA models. 
We also conduct in-depth analysis to examine our method detailedly and provide several hints in the character understanding tasks.

In the future, we plan to apply contrastive learning to other long-form document understanding tasks, such as long document matching~\cite{jiang2019semantic} and fiction understanding~\cite{yu2023personality}.

\section{Limitations}


Our framework depends on pre-trained large languages (PLMs) to encode conversations and summaries, and requires gradient information to tune the PLMs' parameters. This makes it challenging to apply our approach to language models with gigantic sizes. In this work, we demonstrate the generalization of our method in the experimental section at the base and large size, as well as the incapability of ChatGPT-3.5 on character understanding tasks. 
Nevertheless, it remains unclear how well our framework will fit to 3B+ encoder-decoder PLMs or decoder-only LLMs. As our experiments suggest, there is still room for improvement in character understanding tasks.

\bibliography{anthology,custom}

\begin{thebibliography}{58}
\expandafter\ifx\csname natexlab\endcsname\relax\def\natexlab#1{#1}\fi

\bibitem[{Azab et~al.(2019)Azab, Kojima, Deng, and
  Mihalcea}]{azab2019representing}
Mahmoud Azab, Noriyuki Kojima, Jia Deng, and Rada Mihalcea. 2019.
\newblock Representing movie characters in dialogues.
\newblock In \emph{Proceedings of the 23rd Conference on Computational Natural
  Language Learning (CoNLL)}, pages 99--109.

\bibitem[{Bai et~al.(2021)Bai, Zhang, Song, and Xu}]{bai2021joint}
Jiaxin Bai, Hongming Zhang, Yangqiu Song, and Kun Xu. 2021.
\newblock Joint coreference resolution and character linking for multiparty
  conversation.
\newblock \emph{arXiv preprint arXiv:2101.11204}.

\bibitem[{Bamman et~al.(2013)Bamman, O’Connor, and
  Smith}]{bamman2013learning}
David Bamman, Brendan O’Connor, and Noah~A Smith. 2013.
\newblock Learning latent personas of film characters.
\newblock In \emph{Proceedings of the 51st Annual Meeting of the Association
  for Computational Linguistics (Volume 1: Long Papers)}, pages 352--361.

\bibitem[{Beltagy et~al.(2020)Beltagy, Peters, and
  Cohan}]{beltagy2020longformer}
Iz~Beltagy, Matthew~E Peters, and Arman Cohan. 2020.
\newblock Longformer: The long-document transformer.
\newblock \emph{arXiv preprint arXiv:2004.05150}.

\bibitem[{Bower(1978)}]{bower1978experiments}
Gordon~H Bower. 1978.
\newblock Experiments on story comprehension and recall.
\newblock \emph{Discourse Processes}, 1(3):211--231.

\bibitem[{Brahman et~al.(2021)Brahman, Huang, Tafjord, Zhao, Sachan, and
  Chaturvedi}]{brahman2021let}
Faeze Brahman, Meng Huang, Oyvind Tafjord, Chao Zhao, Mrinmaya Sachan, and
  Snigdha Chaturvedi. 2021.
\newblock " let your characters tell their story": A dataset for
  character-centric narrative understanding.
\newblock \emph{arXiv preprint arXiv:2109.05438}.

\bibitem[{Chen et~al.(2021)Chen, Chu, Wiseman, and Gimpel}]{chen2021summscreen}
Mingda Chen, Zewei Chu, Sam Wiseman, and Kevin Gimpel. 2021.
\newblock Summscreen: A dataset for abstractive screenplay summarization.
\newblock \emph{arXiv preprint arXiv:2104.07091}.

\bibitem[{Chen and Choi(2016)}]{chen2016character}
Yu-Hsin Chen and Jinho~D Choi. 2016.
\newblock Character identification on multiparty conversation: Identifying
  mentions of characters in tv shows.
\newblock In \emph{Proceedings of the 17th annual meeting of the special
  interest group on discourse and dialogue}, pages 90--100.

\bibitem[{Currie(2009)}]{currie2009narrative}
Gregory Currie. 2009.
\newblock Narrative and the psychology of character.
\newblock \emph{The journal of aesthetics and art criticism}, 67(1):61--71.

\bibitem[{Dai et~al.(2019)Dai, Yang, Yang, Carbonell, Le, and
  Salakhutdinov}]{dai2019transformer}
Zihang Dai, Zhilin Yang, Yiming Yang, Jaime Carbonell, Quoc~V Le, and Ruslan
  Salakhutdinov. 2019.
\newblock Transformer-xl: Attentive language models beyond a fixed-length
  context.
\newblock \emph{arXiv preprint arXiv:1901.02860}.

\bibitem[{Fiske et~al.(1979)Fiske, Taylor, Etcoff, and
  Laufer}]{fiske1979imaging}
Susan~T Fiske, Shelley~E Taylor, Nancy~L Etcoff, and Jessica~K Laufer. 1979.
\newblock Imaging, empathy, and causal attribution.
\newblock \emph{Journal of Experimental Social Psychology}, 15(4):356--377.

\bibitem[{Flekova and Gurevych(2015)}]{flekova2015personality}
Lucie Flekova and Iryna Gurevych. 2015.
\newblock Personality profiling of fictional characters using sense-level links
  between lexical resources.
\newblock In \emph{Proceedings of the 2015 Conference on Empirical Methods in
  Natural Language Processing}, pages 1805--1816.

\bibitem[{Gao et~al.(2021)Gao, Yao, and Chen}]{gao2021simcse}
Tianyu Gao, Xingcheng Yao, and Danqi Chen. 2021.
\newblock Simcse: Simple contrastive learning of sentence embeddings.
\newblock \emph{arXiv preprint arXiv:2104.08821}.

\bibitem[{Gernsbacher et~al.(1998)Gernsbacher, Hallada, and
  Robertson}]{gernsbacher1998automatically}
Morton~Ann Gernsbacher, Brenda~M Hallada, and Rachel~RW Robertson. 1998.
\newblock How automatically do readers infer fictional characters' emotional
  states?
\newblock \emph{Scientific studies of reading}, 2(3):271--300.

\bibitem[{Guan et~al.(2020)Guan, Huang, Zhao, Zhu, and
  Huang}]{guan2020knowledge}
Jian Guan, Fei Huang, Zhihao Zhao, Xiaoyan Zhu, and Minlie Huang. 2020.
\newblock A knowledge-enhanced pretraining model for commonsense story
  generation.
\newblock \emph{Transactions of the Association for Computational Linguistics},
  8:93--108.

\bibitem[{Han et~al.(2021)Han, Hong, Kim, Ko, and Seo}]{han2021fine}
Janghoon Han, Taesuk Hong, Byoungjae Kim, Youngjoong Ko, and Jungyun Seo. 2021.
\newblock Fine-grained post-training for improving retrieval-based dialogue
  systems.
\newblock In \emph{Proceedings of the 2021 Conference of the North American
  Chapter of the Association for Computational Linguistics: Human Language
  Technologies}, pages 1549--1558.

\bibitem[{Hogan et~al.(2022)Hogan, Li, and Shang}]{hogan2022fine}
William Hogan, Jiacheng Li, and Jingbo Shang. 2022.
\newblock Fine-grained contrastive learning for relation extraction.
\newblock \emph{arXiv preprint arXiv:2205.12491}.

\bibitem[{Inoue et~al.(2022)Inoue, Pethe, Kim, and Skiena}]{inoue2022learning}
Naoya Inoue, Charuta Pethe, Allen Kim, and Steven Skiena. 2022.
\newblock Learning and evaluating character representations in novels.
\newblock In \emph{Findings of the Association for Computational Linguistics:
  ACL 2022}, pages 1008--1019.

\bibitem[{Iyyer et~al.(2016)Iyyer, Guha, Chaturvedi, Boyd-Graber, and
  Daum{\'e}~III}]{iyyer2016feuding}
Mohit Iyyer, Anupam Guha, Snigdha Chaturvedi, Jordan Boyd-Graber, and Hal
  Daum{\'e}~III. 2016.
\newblock Feuding families and former friends: Unsupervised learning for
  dynamic fictional relationships.
\newblock In \emph{Proceedings of the 2016 Conference of the North American
  Chapter of the Association for Computational Linguistics: Human Language
  Technologies}, pages 1534--1544.

\bibitem[{Jiang et~al.(2019)Jiang, Zhang, Li, Bendersky, Golbandi, and
  Najork}]{jiang2019semantic}
Jyun-Yu Jiang, Mingyang Zhang, Cheng Li, Michael Bendersky, Nadav Golbandi, and
  Marc Najork. 2019.
\newblock Semantic text matching for long-form documents.
\newblock In \emph{The world wide web conference}, pages 795--806.

\bibitem[{Joshi et~al.(2020)Joshi, Chen, Liu, Weld, Zettlemoyer, and
  Levy}]{joshi2020spanbert}
Mandar Joshi, Danqi Chen, Yinhan Liu, Daniel~S Weld, Luke Zettlemoyer, and Omer
  Levy. 2020.
\newblock Spanbert: Improving pre-training by representing and predicting
  spans.
\newblock \emph{Transactions of the Association for Computational Linguistics},
  8:64--77.

\bibitem[{Kim and Klinger(2019)}]{kim2019frowning}
Evgeny Kim and Roman Klinger. 2019.
\newblock Frowning frodo, wincing leia, and a seriously great friendship:
  Learning to classify emotional relationships of fictional characters.
\newblock \emph{arXiv preprint arXiv:1903.12453}.

\bibitem[{Kim et~al.(2021)Kim, Yoo, and Lee}]{kim2021self}
Taeuk Kim, Kang~Min Yoo, and Sang-goo Lee. 2021.
\newblock Self-guided contrastive learning for bert sentence representations.
\newblock \emph{arXiv preprint arXiv:2106.07345}.

\bibitem[{Krishnan and Eisenstein(2014)}]{krishnan2014you}
Vinodh Krishnan and Jacob Eisenstein. 2014.
\newblock " you're mr. lebowski, i'm the dude": Inducing address term formality
  in signed social networks.
\newblock \emph{arXiv preprint arXiv:1411.4351}.

\bibitem[{Lee et~al.(2020)Lee, Lee, and Hwang}]{lee2020contrastive}
Seanie Lee, Dong~Bok Lee, and Sung~Ju Hwang. 2020.
\newblock Contrastive learning with adversarial perturbations for conditional
  text generation.
\newblock \emph{arXiv preprint arXiv:2012.07280}.

\bibitem[{Li et~al.(2022)Li, Li, Zhang, Li, Wei, Cui, and Wang}]{li2022c3kg}
Dawei Li, Yanran Li, Jiayi Zhang, Ke~Li, Chen Wei, Jianwei Cui, and Bin Wang.
  2022.
\newblock C3kg: A chinese commonsense conversation knowledge graph.
\newblock \emph{arXiv preprint arXiv:2204.02549}.

\bibitem[{Li et~al.(2020)Li, Yin, Li, Zhang, Hu, Zhang, Wang, Hu, Dong, Wei
  et~al.}]{li2020oscar}
Xiujun Li, Xi~Yin, Chunyuan Li, Pengchuan Zhang, Xiaowei Hu, Lei Zhang, Lijuan
  Wang, Houdong Hu, Li~Dong, Furu Wei, et~al. 2020.
\newblock Oscar: Object-semantics aligned pre-training for vision-language
  tasks.
\newblock In \emph{Computer Vision--ECCV 2020: 16th European Conference,
  Glasgow, UK, August 23--28, 2020, Proceedings, Part XXX 16}, pages 121--137.
  Springer.

\bibitem[{Li and Zhao(2021)}]{li2021self}
Yiyang Li and Hai Zhao. 2021.
\newblock Self-and pseudo-self-supervised prediction of speaker and
  key-utterance for multi-party dialogue reading comprehension.
\newblock \emph{arXiv preprint arXiv:2109.03772}.

\bibitem[{Liu et~al.(2019)Liu, Ott, Goyal, Du, Joshi, Chen, Levy, Lewis,
  Zettlemoyer, and Stoyanov}]{liu2019roberta}
Yinhan Liu, Myle Ott, Naman Goyal, Jingfei Du, Mandar Joshi, Danqi Chen, Omer
  Levy, Mike Lewis, Luke Zettlemoyer, and Veselin Stoyanov. 2019.
\newblock Roberta: A robustly optimized bert pretraining approach.
\newblock \emph{arXiv preprint arXiv:1907.11692}.

\bibitem[{Ma et~al.(2022)Ma, Zhang, and Zhao}]{ma2022structural}
Xinbei Ma, Zhuosheng Zhang, and Hai Zhao. 2022.
\newblock Structural characterization for dialogue disentanglement.
\newblock In \emph{Proceedings of the 60th Annual Meeting of the Association
  for Computational Linguistics (Volume 1: Long Papers)}, pages 285--297.

\bibitem[{Massey et~al.(2015)Massey, Xia, Bamman, and
  Smith}]{massey2015annotating}
Philip Massey, Patrick Xia, David Bamman, and Noah~A Smith. 2015.
\newblock Annotating character relationships in literary texts.
\newblock \emph{arXiv preprint arXiv:1512.00728}.

\bibitem[{McKee(1997)}]{mckee1997story}
Robert McKee. 1997.
\newblock \emph{Story: style, structure, substance, and the principles of
  screenwriting}.
\newblock Harper Collins.

\bibitem[{Mead(1990)}]{mead1990representation}
Gerald Mead. 1990.
\newblock The representation of fictional character.
\newblock \emph{Style}, pages 440--452.

\bibitem[{Mikolov et~al.(2013)Mikolov, Sutskever, Chen, Corrado, and
  Dean}]{mikolov2013distributed}
Tomas Mikolov, Ilya Sutskever, Kai Chen, Greg~S Corrado, and Jeff Dean. 2013.
\newblock Distributed representations of words and phrases and their
  compositionality.
\newblock \emph{Advances in neural information processing systems}, 26.

\bibitem[{Min et~al.(2021)Min, Ross, Sulem, Veyseh, Nguyen, Sainz, Agirre,
  Heinz, and Roth}]{min2021recent}
Bonan Min, Hayley Ross, Elior Sulem, Amir Pouran~Ben Veyseh, Thien~Huu Nguyen,
  Oscar Sainz, Eneko Agirre, Ilana Heinz, and Dan Roth. 2021.
\newblock Recent advances in natural language processing via large pre-trained
  language models: A survey.
\newblock \emph{arXiv preprint arXiv:2111.01243}.

\bibitem[{Mostafazadeh et~al.(2016)Mostafazadeh, Chambers, He, Parikh, Batra,
  Vanderwende, Kohli, and Allen}]{mostafazadeh2016corpus}
Nasrin Mostafazadeh, Nathanael Chambers, Xiaodong He, Devi Parikh, Dhruv Batra,
  Lucy Vanderwende, Pushmeet Kohli, and James Allen. 2016.
\newblock A corpus and cloze evaluation for deeper understanding of commonsense
  stories.
\newblock In \emph{Proceedings of the 2016 Conference of the North American
  Chapter of the Association for Computational Linguistics: Human Language
  Technologies}, pages 839--849.

\bibitem[{Onions et~al.(1966)Onions, Friedrichsen, Burchfield
  et~al.}]{onions1966oxford}
Charles~Talbut Onions, George Washington~Salisbury Friedrichsen, Robert~William
  Burchfield, et~al. 1966.
\newblock \emph{The Oxford dictionary of English etymology}, volume 178.
\newblock Clarendon Press Oxford.

\bibitem[{Pan et~al.(2021)Pan, Wang, Wu, and Li}]{pan2021contrastive}
Xiao Pan, Mingxuan Wang, Liwei Wu, and Lei Li. 2021.
\newblock Contrastive learning for many-to-many multilingual neural machine
  translation.
\newblock \emph{arXiv preprint arXiv:2105.09501}.

\bibitem[{Paris and Paris(2003)}]{paris2003assessing}
Alison~H Paris and Scott~G Paris. 2003.
\newblock Assessing narrative comprehension in young children.
\newblock \emph{Reading Research Quarterly}, 38(1):36--76.

\bibitem[{Pradhan et~al.(2012)Pradhan, Moschitti, Xue, Uryupina, and
  Zhang}]{pradhan2012conll}
Sameer Pradhan, Alessandro Moschitti, Nianwen Xue, Olga Uryupina, and Yuchen
  Zhang. 2012.
\newblock Conll-2012 shared task: Modeling multilingual unrestricted
  coreference in ontonotes.
\newblock In \emph{Joint conference on EMNLP and CoNLL-shared task}, pages
  1--40.

\bibitem[{Qin et~al.(2022)Qin, Chen, Xie, Li, Lou, Che, and Kan}]{qin2022gl}
Libo Qin, Qiguang Chen, Tianbao Xie, Qixin Li, Jian-Guang Lou, Wanxiang Che,
  and Min-Yen Kan. 2022.
\newblock Gl-clef: A global-local contrastive learning framework for
  cross-lingual spoken language understanding.
\newblock \emph{arXiv preprint arXiv:2204.08325}.

\bibitem[{Qin et~al.(2020)Qin, Lin, Takanobu, Liu, Li, Ji, Huang, Sun, and
  Zhou}]{qin2020erica}
Yujia Qin, Yankai Lin, Ryuichi Takanobu, Zhiyuan Liu, Peng Li, Heng Ji, Minlie
  Huang, Maosong Sun, and Jie Zhou. 2020.
\newblock Erica: improving entity and relation understanding for pre-trained
  language models via contrastive learning.
\newblock \emph{arXiv preprint arXiv:2012.15022}.

\bibitem[{Qiu et~al.(2020)Qiu, Sun, Xu, Shao, Dai, and Huang}]{qiu2020pre}
Xipeng Qiu, Tianxiang Sun, Yige Xu, Yunfan Shao, Ning Dai, and Xuanjing Huang.
  2020.
\newblock Pre-trained models for natural language processing: A survey.
\newblock \emph{Science China Technological Sciences}, 63(10):1872--1897.

\bibitem[{Radford et~al.(2021)Radford, Kim, Hallacy, Ramesh, Goh, Agarwal,
  Sastry, Askell, Mishkin, Clark et~al.}]{radford2021learning}
Alec Radford, Jong~Wook Kim, Chris Hallacy, Aditya Ramesh, Gabriel Goh,
  Sandhini Agarwal, Girish Sastry, Amanda Askell, Pamela Mishkin, Jack Clark,
  et~al. 2021.
\newblock Learning transferable visual models from natural language
  supervision.
\newblock In \emph{International conference on machine learning}, pages
  8748--8763. PMLR.

\bibitem[{Saha(2021)}]{Saha_Movie_Script_Database_2021}
Aveek Saha. 2021.
\newblock \href {https://github.com/Aveek-Saha/Movie-Script-Database} {{Movie
  Script Database}}.

\bibitem[{Sang et~al.(2022{\natexlab{a}})Sang, Mou, Li, Stanton, and
  Yu}]{sang2022survey}
Yisi Sang, Xiangyang Mou, Jing Li, Jeffrey Stanton, and Mo~Yu.
  2022{\natexlab{a}}.
\newblock A survey of machine narrative reading comprehension assessments.
\newblock \emph{arXiv preprint arXiv:2205.00299}.

\bibitem[{Sang et~al.(2022{\natexlab{b}})Sang, Mou, Yu, Yao, Li, and
  Stanton}]{sang2022tvshowguess}
Yisi Sang, Xiangyang Mou, Mo~Yu, Shunyu Yao, Jing Li, and Jeffrey Stanton.
  2022{\natexlab{b}}.
\newblock Tvshowguess: Character comprehension in stories as speaker guessing.
\newblock \emph{arXiv preprint arXiv:2204.07721}.

\bibitem[{Shu et~al.(2021)Shu, Zhang, Dong, Shi, Yu, and Zhang}]{shu2021logic}
Chang Shu, Yusen Zhang, Xiangyu Dong, Peng Shi, Tao Yu, and Rui Zhang. 2021.
\newblock Logic-consistency text generation from semantic parses.
\newblock \emph{arXiv preprint arXiv:2108.00577}.

\bibitem[{Tu et~al.(2022)Tu, Li, Cui, Wang, Wen, and Yan}]{tu2022misc}
Quan Tu, Yanran Li, Jianwei Cui, Bin Wang, Ji-Rong Wen, and Rui Yan. 2022.
\newblock Misc: A mixed strategy-aware model integrating comet for emotional
  support conversation.
\newblock \emph{arXiv preprint arXiv:2203.13560}.

\bibitem[{Vamvas and Sennrich(2021)}]{vamvas2021contrastive}
Jannis Vamvas and Rico Sennrich. 2021.
\newblock Contrastive conditioning for assessing disambiguation in mt: A case
  study of distilled bias.

\bibitem[{Van~der Maaten and Hinton(2008)}]{van2008visualizing}
Laurens Van~der Maaten and Geoffrey Hinton. 2008.
\newblock Visualizing data using t-sne.
\newblock \emph{Journal of machine learning research}, 9(11).

\bibitem[{Yang et~al.(2022)Yang, Li, Zhang, Xiao, Liu, Yuan, and
  Gao}]{yang2022unified}
Jianwei Yang, Chunyuan Li, Pengchuan Zhang, Bin Xiao, Ce~Liu, Lu~Yuan, and
  Jianfeng Gao. 2022.
\newblock Unified contrastive learning in image-text-label space.
\newblock In \emph{Proceedings of the IEEE/CVF Conference on Computer Vision
  and Pattern Recognition}, pages 19163--19173.

\bibitem[{Yao et~al.(2023)Yao, Peng, Mao, and Luo}]{yao2023exploring}
Liang Yao, Jiazhen Peng, Chengsheng Mao, and Yuan Luo. 2023.
\newblock Exploring large language models for knowledge graph completion.
\newblock \emph{arXiv preprint arXiv:2308.13916}.

\bibitem[{Yu et~al.(2023)Yu, Li, Yao, Pang, Zhou, Xiao, Meng, and
  Zhou}]{yu2023personality}
Mo~Yu, Jiangnan Li, Shunyu Yao, Wenjie Pang, Xiaochen Zhou, Zhou Xiao, Fandong
  Meng, and Jie Zhou. 2023.
\newblock Personality understanding of fictional characters during book
  reading.
\newblock \emph{arXiv preprint arXiv:2305.10156}.

\bibitem[{Zaheer et~al.(2020)Zaheer, Guruganesh, Dubey, Ainslie, Alberti,
  Ontanon, Pham, Ravula, Wang, Yang et~al.}]{zaheer2020big}
Manzil Zaheer, Guru Guruganesh, Kumar~Avinava Dubey, Joshua Ainslie, Chris
  Alberti, Santiago Ontanon, Philip Pham, Anirudh Ravula, Qifan Wang, Li~Yang,
  et~al. 2020.
\newblock Big bird: Transformers for longer sequences.
\newblock \emph{Advances in neural information processing systems},
  33:17283--17297.

\bibitem[{Zhang et~al.(2022{\natexlab{a}})Zhang, Li, Yang, and
  Li}]{zhang2022fine}
Hengyuan Zhang, Dawei Li, Shiping Yang, and Yanran Li. 2022{\natexlab{a}}.
\newblock Fine-grained contrastive learning for definition generation.
\newblock \emph{arXiv preprint arXiv:2210.00543}.

\bibitem[{Zhang et~al.(2022{\natexlab{b}})Zhang, Ji, Zhang, and
  Passonneau}]{zhang2022contrastive}
Rui Zhang, Yangfeng Ji, Yue Zhang, and Rebecca~J Passonneau.
  2022{\natexlab{b}}.
\newblock Contrastive data and learning for natural language processing.
\newblock In \emph{Proceedings of the 2022 Conference of the North American
  Chapter of the Association for Computational Linguistics: Human Language
  Technologies: Tutorial Abstracts}, pages 39--47.

\bibitem[{Zhou and Choi(2018)}]{zhou2018they}
Ethan Zhou and Jinho~D Choi. 2018.
\newblock They exist! introducing plural mentions to coreference resolution and
  entity linking.
\newblock In \emph{Proceedings of the 27th International Conference on
  Computational Linguistics}, pages 24--34.

\end{thebibliography}
\bibliographystyle{acl_natbib}

\appendix
\onecolumn

\section{Example of Script-based Character Understanding Task}
\label{Example of Script-based Character Understanding Task}

\begin{table*}[h]
\centering
\begin{tabular}{lp{4.5in}p{2.9in}}
\hline
Input & P0: Hey, sorry about that P1: No, we're sorry. We never should have been comparing relationships in the first place. P2: Why? We won. You know, I say, next, we take on Koothrappali and his dog. Really give ourselves a challenge. P3: I just want to say one more thing about this. Just because Penny and I are very different people does not mean that we're a bad couple. P2: The answer is one simple test away. Hmm? You know, it's like when I thought there was a possum in my closet. Did I sit around wondering? No, I sent Leonard in with a pointy stick and a bag. P3: I killed his Chewbacca slippers. P0: Let's just take the test. P3: No, no, no, I don't want to. P0: Oh, well, 'cause you know we're gonna do bad. P3: Because it doesn't matter. I don't care if we're a ten or a two. P2: Or a one. A one is possible. P3: Marriage is scary. You're scared, I'm scared. But it doesn't make me not want to do it. It, it just makes me want to hold your hand and do it with you. P0: Leonard. P1: It makes me so happy if you said things like that. P2: We got an eight-point-two. Trust me, you're happy. \\ \hline
Label & P0: Penny, P1: Amy, P2: Sheldon, P3: Leonard                                                                                                                                                                                                                                                                                                                                                                                                                                                                                                                                                                                                                                                                                                                                                                                                                                                                                                                                                                                                                                                                                          \\ \hline
\end{tabular}
\caption{A example from character guessing task.}
\label{example: character guessing}
\end{table*}

\begin{table*}[h]
\centering
\begin{tabular}{lp{4.5in}p{2.9in}}
\hline
\multicolumn{1}{l}{Input}    & Ross: I told mom and dad last night, they seemed to take it pretty well. Monica: Oh really, so that hysterical phone call I got from a woman at sobbing 3:00 A.M., "I'll never have grandchildren, I'll never have grandchildren." was what? A wrong number? Ross: Sorry. Joey: Alright Ross, look. You're feeling a lot of pain right now. You're angry. You're hurting. Can I tell you what the answer is? \\ \hline
\multicolumn{1}{l}{Label-CR} & Ross: \textcolor{red}{I} told \textcolor{blue}{mom} and \textcolor{purple}{dad} last night, they seemed to take it pretty well. Monica: Oh really, so that hysterical phone call \textcolor{brown}{I} got from a \textcolor{blue}{woman} at sobbing 3:00 A.M., "\textcolor{blue}{I}'ll never have grandchildren, \textcolor{blue}{I}'ll never have grandchildren." was what? A wrong number? Ross: Sorry. Joey: Alright \textcolor{red}{Ross}, look. \textcolor{red}{You}'re feeling a lot of pain right now. \textcolor{red}{You}'re angry. \textcolor{red}{You}'re hurting. Can \textcolor{green}{I} tell \textcolor{red}{you} what the answer is? \\ \hline
Label-CL                       & I: Ross Geller, mom: Judy Geller, dad: Jack Geller, I: Monica Geller, woman: Judy Geller, I: Monica Geller, I: Monica Geller, Ross: Ross Geller, You: Ross Geller, You: Ross Geller, You: Ross Geller, I: Joey Tribbiani, you: Ross Geller                
\\ \hline
\end{tabular}
\caption{A example from coreference resolution and character linking tasks. For coreference resolution, the goal of the task is to cluster the coreferences that refer to the same character in one group (we use the same color to represent).}
\label{example: coreference resolution and character linking}
\end{table*}

\section{Details of Character Embedding Generation}
\label{Details of Character Embedding Generation}
Here we give detailed formulations of our character embedding extraction and character-level attention process.

\noindent\textbf{Coreference Resolution \& Character Linking}
Given the context encoding $\mathbf{H}$, we follow~\cite{bai2021joint} to initialize the mention-level character embedding:
\begin{equation}
    e_i = t_{start_i} + t_{end_i} + e_{speaker_i}
\end{equation}
where $t_{start_i}$ and $t_{end_i}$ are the contextualized representation of the beginning and the end tokens of mention $i$, and the $e_{speaker_i}$ is the speaker embedding for the current speaker of the utterance where the $i_{th}$ mention belong to. The speaker embeddings are randomly initialized before training.

\noindent\textbf{Character Guessing}
We follow~\cite{sang2022tvshowguess} to extract speaker-level character embedding from the context encoding $\mathbf{H}$:
\begin{equation}
    A = {\rm Attention}(\mathbf{H})
\end{equation}
\begin{equation}
    a_i = {\rm Softmax}(A \odot M_i)
\end{equation}
\begin{equation}
    e_i = \mathbf{H}^Ta_i
\end{equation}

where Attention($\cdot$) is a one-layer feedforward network to compute the token-level attention weight. $M_i$ is a token-level mask such that $M_i[j]=1$ if the $j_{th}$ word belongs to an utterance of the $i_{th}$ anonymous speaker and $Mx[j]=0$ otherwise. $a_i$ is the token weight used to pool the hidden states to summarize a character representation.

After obtaining character embedding, we adopt the MLSA layer we mentioned in Section \ref{Character Representation in Summary} to gather information for each character embedding:
\begin{equation}
    e_1, ...,e_n = {\rm MLSA}(e_1, ...,e_n)
\end{equation}

\section{Prompts for ChatGPT}
\label{Prompt for ChatGPT}

For character linking, as Table~\ref{Prompt Character Linking} shows, we provide the original scripts' content for ChatGPT, followed by the position of the mention to be inferenced and all the optional characters. 
For coreference resolution, we tried several different prompts to ask ChatGPT to do clustering. 
However, there is always omitting of mentions\footnote{For example, there are 20 mentions to be clustered in a sample but the model's output just contains the clustering result of 15 of them.} in the models' output which leads to very poor performance on coreference resolution. 
So we just use the model's output on character linking as the clustering results of each character and calculate the corresponding metrics for coreference resolution.

For character guessing, we provide the show name of the scripts and optional characters to the model. We concatenate them in front of the script's content and input the prompt to ChatGPT to do inference as Table~\ref{Prompt Character Guessing} shows. 
For zero-shot, we try asking ChatGPT to guess one character in each request (represent as Prompt-Character) and guess all characters in each request (represent as Prompt-Sample) and find the latter performs better as Table~\ref{Prompt Comparision} shows.
One possible reason for that is the model would attend to more information from other characters' utterances given Prompt-Sample, which is exactly the key to perform well in character guessing.
For one-shot, we only adopt the Prompt-Sample due to its superior performance in the zero-shot setting.

For the one-shot setting, we additionally provide the model a sample together with its label as a demonstration.
For both tasks, after getting the output of the model, we also use the RE module provided by Python to map the raw text to the most similar label.

\begin{table}[h!]\small
\centering
\begin{tabular}{c|cc}
\hline
                 & \multicolumn{1}{c}{MICRO} & \multicolumn{1}{c}{MACRO} \\ \hline
Prompt-Character &  43.77                    & 37.87                      \\ 
Prompt-Sample    & \textbf{48.58}                      & \textbf{42.17}                      \\ \hline
\end{tabular}
\caption{Comparison of the two prompt methods in character guessing with zero-shot.}
\label{Prompt Comparision}
\end{table}

\section{Detailed Training Settings}
\label{Detailed Training Settings}

For every available positive pair \footnote{Excluding the situation that a character show in the dialogue but not in the summary, and vice versa.} of the same character, we randomly choose one to conduct contrastive learning. 
We set the $\lambda$, $\alpha$ and $\beta$ to 1.0, 0.5, 0.5 respectively in the first stage of training for all three tasks. 
We test SpanBERT and C$^2$ in coreference resolution and character linking and Longformer-P and BigBird-P in character guessing. 
We use ChatGPT in all three tasks. 
Table~\ref{other hyper-parameter settings} gives parameter settings of the learning rate, batch size, and training epochs in the two stages of learning. 
We use Pytorch-1.8.1 deep learning framework and Transformers-4.1.2 library for our experiment. 
We train our models on a single A40 GPU.
It takes about 3 hours to train a SpanBert-base/ C$^2$-base model and 20 hours to train a Longformer-P-base/ BigBird-P-base model. 
Large-version model training takes twice the time. 
Table~\ref{datasets} shows the detailed statistics of the datasets we use.
We also give detailed information about models in the base and large size we use in the experiment in Table~\ref{model details}.

\begin{table*}[h!]\small
\centering
\begin{tabular}{c|ccc|ccc}
\hline
\multirow{2}{*}{Dataset}                                                               & \multicolumn{3}{c|}{First Stage}                                    & \multicolumn{3}{c}{Second Stage}                                   \\ \cline{2-7} 
                                                                                       & \multicolumn{1}{c|}{LR}   & \multicolumn{1}{c|}{Batch Size} & Epoch & \multicolumn{1}{c|}{LR}   & \multicolumn{1}{c|}{Batch Size} & Epoch \\ \hline
\begin{tabular}[c]{@{}c@{}}Coreference Resolution \& \\ Character Linking\end{tabular} & \multicolumn{1}{c|}{1e-5} & \multicolumn{1}{c|}{-}          & 30    & \multicolumn{1}{c|}{2e-5} & \multicolumn{1}{c|}{-}          & 100   \\ \hline
Character Guessing                                                                     & \multicolumn{1}{c|}{4e-6} & \multicolumn{1}{c|}{4}          & 20    & \multicolumn{1}{c|}{2e-5} & \multicolumn{1}{c|}{2}          & 40    \\ \hline
\end{tabular}
\caption{Other hyper-parameters settings in our experiments. Note that for coreference resolution and character linking, we follow the previous works~\cite{chen2016character,bai2021joint} and incorporate every sample inside a scene in a batch.}
\label{other hyper-parameter settings}
\end{table*}

\begin{table}[h!]\small
\centering
\begin{tabular}{c|cccc}
\hline
Task                                                                                  & Train & Validation & Test & Total \\ \hline
\begin{tabular}[c]{@{}c@{}}Character Linking \&\\ Coreference Resolution\end{tabular} & 987   & 122        & 192  & 1301  \\ \hline
Character Guessing                                                                    & 10071 & 819        & 823  & 11713 \\ \hline
\end{tabular}
\caption{Detailed information about the datasets for each task.}
\label{datasets}
\end{table}

\begin{table}[]\small
\centering
\begin{tabular}{c|cccc}
\hline
Size  & Parameter Size & \begin{tabular}[c]{@{}c@{}}Number of\\ Transformer Layer\end{tabular} & \begin{tabular}[c]{@{}c@{}}Number of \\ Attention Head\end{tabular} & Hidden Size \\ \hline
Base  & 125M           & 12                                                                    & 12                                                                  & 768         \\ \hline
Large & 354M           & 24                                                                    & 16                                                                  & 1024        \\ \hline
\end{tabular}
\caption{Detailed information about models in the base and large size we use in the experiment.}
\label{model details}
\end{table}

\begin{table}[t!]\small
\centering
\begin{tabular}{@{}cl@{}}
\toprule
Scripts & \begin{tabular}[c]{@{}l@{}}Monica Geller: Tell him. \\ Rachel Green: No. \\ Phoebe Buffay: Tell him, tell him. \\ Monica Geller: Just...please tell him. \\ Rachel Green: Shut up! \\ Chandler Bing: Tell me what? \\ Monica Geller: Look at you, you won't even look at him . \\ Chandler Bing: Oh, come on tell me. I could use another reason why women won't \\ look at me. \\ Rachel Green: All right, all right, all right. Last night, I had a dream that, uh, \\ you and I, were... \\ Phoebe Buffay: Doing it on this table . \\ Chandler Bing: Wow ! \\ Joey Tribbiani: Exellent dream score . \\ Ross Geller: Why, why, why would you dream that? \\ Chandler Bing: More importantly, was I any good? \\ Rachel Green: Well, you were pretty damn good. \\ Chandler Bing: Interesting, cause in my dreams, I'm allways surprisingly inadequate. \\ Rachel Green: Well, last night you seemed to know your way around the table. \\ Ross Geller: I love it, when we share. Chandler Bing: You're okay there? \\ Ross Geller: I can't believe you two had sex in her dream. \\ Chandler Bing: I'm sorry, it was a one-time - thing. I was very drunk and i was somebody\\ else's subconscious .\end{tabular} \\ \midrule
Prompt  & \begin{tabular}[c]{@{}l@{}} \textcolor{red}{Here is an example:} \\ \textcolor{red}{<Demonstration>} \\ \textcolor{red}{Following this example and} read the following conversation: \\ \textless{}Scripts\textgreater\\ The NO.1 "him" in the utterance "Tell him ." refer to which character? \\ you should choose answer from Ross Geller, Rachel Green, Chandler Bing, \\ Monica Geller, Joey Tribbiani, Phoebe Buffay, Emily, Richard Burke, Carol Willick, \\ Ben Geller, Peter Becker, Judy Geller, Barry Farber, Jack Geller, Kate Miller, \\ \#OTHER\#, \#GENERAL\#\end{tabular} \\ \bottomrule
\end{tabular}
\caption{Original scripts and the prompt we input to ChatGPT in coreference resolution and character linking. We replace the original scripts in the prompt with a special token <Scripts> to save space. We use \textcolor{red}{red} color to represent the additional prompt and demonstration adopted in one-shot settings.}
\label{Prompt Character Linking}
\end{table}

\begin{table}[t!]\small
\centering
\begin{tabular}{@{}cl@{}}
\toprule
\multicolumn{1}{c}{Scripts} & \begin{tabular}[c]{@{}l@{}}P0 : hey, frasier, i'm glad i caught you. did you just get home? \\ P1 : no, i've been here a while. can't bring myself to go in. not with her in there. \\ P1 : she's getting better. \\ P0 : look, i did you a favor. my lawyer drew up this document, it releases you \\ from all liability, if you just get ann to sign it. \\ P1 : oh, roz, there's no way i'm going to get her to sign this. but i have a better \\ plan: i've just booked passage for her and her mother on a two week cruise to \\ alaska. that way i'll get her out of my home, but we'll still feel like we're friends. \\ P0 : hmm, not a bad idea. good luck with that. \\ P1 : thank you. roz, i've been meaning to ask you: how did you ever become \\ friends with ann? i mean, she's really not your type, is she? \\ P0 : oh, we're not really friends. i rear ended her in 1989. P1 : oh, the tickets \\ arrived. P1 : well, i'd hoped she would be. P1 : you told her i was taking you? \\ P1 : ann... \\ P1 : hold that thought, while bunny goes and pours himself a big ol' glass o' \\ wine. P1 : caroline. \\ P1 : uh... \\ P1 : uh...just a neighbor. \\ P1 : {[}pained{]} no. since we made our plans, caroline, i've met someone else. \\ P1 : just go! \\ P1 : {[}darkly{]} i don't know what i'm gonna do with you either. \\ P1 : oh, it's nothing. just some work stuff. \\ P1 : no. \\ P1 : no! \\ P1 : no, never! \\ P1 : i realize that you're angry now, bunny... \\ P1 : all right, fine! go ahead and sue! i am fed up with this charade! this was an \\ accident! i have cared for you, i have waited on you, i have pumiced your heels \\ and set your hair! well, if that's not enough for you, so be it! i don't care \\ anymore , i will not beg! you can take me to the cleaners but you cannot take \\ my dignity! P1 : oh dear god, please, no! pleas, no, no, please! please, please \\ don't  sue me! my...things, my beautiful, beautiful things. i love them so... \\ P1 : {[}weeping{]} no. \\ P1 : you will? \\ P1 : thank you, ann. i'm sorry it had to come down to all this, this legal business. \\ if it were up to me, i would tear up this piece of paper and forget everything that's \\ happened here. P1 : and, uh, here. \\ P1 : and...here.\end{tabular} \\ \midrule
Prompt-Character                     & \begin{tabular}[c]{@{}l@{}}\textcolor{red}{Here is an example:} \\ \textcolor{red}{<Demonstration>} \\ \textcolor{red}{Following this example and} tell me P0 below are which character from TV show "Frasier", \\ please choose from frasier, roz, niles, martin, daphne and bob: \\<Scripts> \\
Tell me P1 below are which character from TV show "Frasier", \\ please choose from frasier, roz, niles, martin, daphne and bob: \\<Scripts>
\end{tabular} \\
\midrule
Prompt-Sample                      & \begin{tabular}[c]{@{}l@{}} \textcolor{red}{Here is an example:} \\ \textcolor{red}{<Demonstration>} \\ \textcolor{red}{Following this example and} tell me P0, P1 below are which character from TV show "Frasier", \\ please choose from frasier, roz, niles, martin, daphne and bob: \\<Scripts>\end{tabular} \\
\bottomrule
\end{tabular}
\caption{Original scripts and the prompt we input to ChatGPT in character guessing. We replace the original scripts in the prompt with a special token <Scripts> to save space. We use \textcolor{red}{red} color to represent the additional prompt and demonstration adopted in one-shot settings.}
\label{Prompt Character Guessing}
\end{table}

\clearpage

\section{Evidence Type Merging}
\label{Evidence Type Merging}
In~\citet{sang2022tvshowguess}, the annotations divide the evidence types of guessing characters into extremely detailed categories, and several psychologists are asked to assign a category of evidence to each character. In specific, there are 9 types of evidence totally for guessing character identification according to~\citet{sang2022tvshowguess}. 
They are attribute, relation, status,	background,	exclusion, mention, linguistic, memory, and personality. 

However, one drawback of the subdivided categories is the scarcity of a certain categories. To address the high variance issue caused by the scarcity, we merge the fine-grained annotations into broader ones. Based on the definition of them in the original paper\footnote{Please refer to the original paper for more details.}, we split them into 2 big categories as Table~\ref{Merged Category} shows. 
Here Global \& In-depth evidence means cases in which the character can only be predicted according to his/her global information (like one's relationship with others) or some subtle clues (like one's Linguistic style). 
Local \& Textual evidence means cases in which a character can be easily predicted only using the content in the local sample (like background information appearing in other characters' utterances) or something very direct (like calling one character's name directly). 
Note that we abandon the evidence exclusion because it is more like a guessing technique rather than evidence.

\begin{table}[h!]\small
\centering
\begin{tabular}{@{}cc@{}}
\toprule
Merged Category                                                    & Original Category                \\ \midrule
\multicolumn{1}{c}{\multirow{6}{*}{Global \& In-depth evidence}} & \multicolumn{1}{c}{Attribute}   \\ \cmidrule(l){2-2} 
\multicolumn{1}{c}{}                                             & \multicolumn{1}{c}{Relation}    \\ \cmidrule(l){2-2} 
\multicolumn{1}{c}{}                                             & \multicolumn{1}{c}{Status}      \\ \cmidrule(l){2-2} 

\multicolumn{1}{c}{}                                             & \multicolumn{1}{c}{Linguistics} \\ \cmidrule(l){2-2} 
\multicolumn{1}{c}{}                                             & \multicolumn{1}{c}{Memory}      \\ \cmidrule(l){2-2} 
\multicolumn{1}{c}{}                                             & \multicolumn{1}{c}{Personality} \\ \midrule
\multicolumn{1}{c}{\multirow{2}{*}{Local \& Textual evidence}}   & \multicolumn{1}{c}{Background}  \\ \cmidrule(l){2-2} 
\multicolumn{1}{c}{}                                             & Mention                          \\ \bottomrule
\end{tabular}
\caption{Merged result of the subdivided evidence type.}
\label{Merged Category}
\end{table}

\end{document}